\documentclass[10pt,twocolumn,letterpaper]{article}

\usepackage[pagenumbers]{cvpr} 

\usepackage{microtype}
\usepackage{amsmath}
\usepackage[table]{xcolor}
\usepackage{comment}
\usepackage{multirow}
\usepackage{makecell}
\usepackage{comment}

\usepackage{microtype}

\definecolor{cvprblue}{rgb}{0.21,0.49,0.74}

\usepackage{hyperref}
\hypersetup{pagebackref,breaklinks,colorlinks,allcolors=cvprblue}

\title{From Concepts to Judgments: Interpretable Image Aesthetic Assessment}

\author{
Xiao-Chang Liu \quad Johan Wagemans\\
Research Unit Brain and Cognition, KU Leuven\\
Leuven, Belgium\\
{\tt\small xiaochang.liu@kuleuven.be, johan.wagemans@kuleuven.be}
}

\begin{document}

\maketitle
\begin{abstract}
  Image aesthetic assessment (IAA) aims to predict the aesthetic quality of images as perceived by humans. While recent IAA models achieve strong predictive performance, they offer little insight into the factors driving their predictions. Yet for users, understanding why an image is considered pleasing or not is as valuable as the score itself, motivating growing interest in interpretability within IAA. When humans evaluate aesthetics, they naturally rely on high-level cues to justify their judgments. Motivated by this observation, we propose an interpretable IAA framework grounded in human-understandable aesthetic concepts. We learn these concepts in an accessible manner, constructing a subspace that forms the foundation of an inherently interpretable model. To capture nuanced influences on aesthetic perception beyond explicit concepts, we introduce a simple yet effective residual predictor. Experiments on photographic and artistic datasets demonstrate that our method achieves competitive predictive performance while offering transparent, human-understandable aesthetic judgments.
\end{abstract}
    
\section{Introduction}
\label{sec:intro}

Image Aesthetic Assessment (IAA) is a long-standing research area within computational aesthetics. As an interdisciplinary field, it bridges science and art, with applications in photography, visual design, and many other fields. A central goal of IAA research is to develop computational models that can automatically assess the aesthetic quality of visual inputs. However, to obtain state-of-the-art performance, most recent models sacrifice human explainability. In this paper, we revisit the IAA with a focus on interpretability: aiming not only to accurately predict aesthetic quality, but to provide clear interpretations for the model's decisions.

Decades ago, approaches to image quality and aesthetic assessment relied on hand-crafted features derived from human expertise and intuition.
These included, but were not limited to, low-level image statistics
(\eg, color~\cite{hasler2003measuring,nishiyama2011aesthetic}, edges~\cite{ke2006design}, luminance~\cite{obrador2009low}),
global composition features~\cite{obrador2010role},
saliency-based cues~\cite{luo2008photo,wong2009saliency}, and content attributes~\cite{you2009perceptual,jiang2010automatic,dhar2011high}, among others.
Nowadays, with the widespread adoption of Deep Neural Networks (DNN), these manually designed features have been largely replaced by automatically learned representations.
This shift has not only alleviated the burden of feature engineering but also led to significant performance gains, establishing DNN as the dominant approach in modern image assessment models~\cite{lu2015deep,bosse2017deep,talebi2018nima}.
However, the opaque nature of DNN makes their decision-making processes difficult to explain. This has led to a growing interest in developing more transparent and interpretable IAA systems.

Consider a scenario in which photographers or designers use a model to assess image aesthetics. They may wonder: Why was this photo rated poorly? Was it due to poor symmetry? Harsh lighting? An unbalanced layout? Unfortunately, most existing models cannot support such queries, as they map directly from pixels to a score. While some models predict intermediate attribute values, the relationship between these values and the final score remains opaque. As a result, these models offer limited insight into the factors driving their predictions, constraining their usefulness in professional contexts.

In this paper, we present an interpretable framework for IAA. Our goal is to build a model whose decisions can be readily understood and interpreted by humans. To this end, we learn a set of high-level aesthetic concepts and use them as the basis for prediction. Specifically, we learn a concept subspace in which each axis corresponds to a meaningful aesthetic property. Each image is then projected into this subspace, and aesthetic predictions are made using a sparse, inherently interpretable linear model.
This linear integration of human-understandable concepts is inspired by empirical studies in aesthetics:
Iigaya~\etal~\cite{iigaya2021aesthetic} showed that human aesthetic preferences can be well approximated by a linear feature summation (LFS) model, and their subsequent neuroimaging work further linked this model to neural mechanisms underlying aesthetic valuation~\cite{iigaya2023neural}.

To construct the concept subspace, we leverage concept-related image sets consisting of positive and negative examples for each aesthetic concept. For each concept, we train a linear classifier and take the normal vector to its decision boundary as the concept activation vector. These vectors span the aesthetic concept space. This design makes our framework particularly well suited to aesthetics, where defining concepts through examples aligns naturally with human perception. While the concept subspace provides strong interpretability, human aesthetic judgments are often influenced by subtle factors that go beyond explicit aesthetic concepts. To account for this inherent complexity, we introduce a residual predictor that complements the concept subspace, modeling additional variance without compromising the interpretable core.

\noindent\textbf{Contributions:} 
We introduce a new modeling perspective for IAA that integrates human-understandable aesthetic concepts in an inherently interpretable model. Instead of relying on post-hoc explanations of black-box predictors, our method performs prediction in a learned concept space through a sparse linear model, ensuring that the decision process is transparent by design. To account for aspects of aesthetic judgment that may not be fully captured by explicit concepts, we incorporate a lightweight residual predictor that complements the interpretable core while preserving its integrity. We evaluate the framework across several benchmark datasets covering both photography and art domains, demonstrating its feasibility and competitive performance.

\section{Related Work}
\label{sec:relatedwork}

The scientific study of human aesthetics has a long history. In psychology and cognitive science, empirical aesthetics has explored how people perceive and evaluate visual beauty~\cite{leder2004model,palmer2013visual,pelowski2017move}. One of the earliest quantitative theories was proposed by Birkhoff in \textit{Aesthetic Measure} (1933)~\cite{birkhoff1933aesthetic}, which is widely regarded as the origin of computational aesthetics.
Birkhoff proposed a simple formula: $M=O/C$, where $O$ denotes order, $C$ complexity, and $M$ the resulting aesthetic value. 
Over the past couple of decades, a wide range of techniques have been developed~\cite{bo2018computational,anwar2022image,daryanavard2025deep}. These include both handcrafted features-based and modern deep learning-based approaches. Our work is motivated by four observations, from which we developed our contributions.

\noindent\textbf{Observation 1: Traditional IAA methods are interpretable, but primarily rely on manually designed feature representations.} 
Early approaches to Image Aesthetic Assessment (IAA) relied heavily on manually designed features informed by human intuition and domain expertise.
These methods typically encoded aesthetic attributes using handcrafted filters, and employed classical machine learning algorithms for prediction.

Standard features include color~\cite{nishiyama2011aesthetic}, contrast~\cite{aydin2014automated}, composition~\cite{li2010towards,liu2010optimizing}, texture~\cite{lo2012assessment} and foreground-background statistics~\cite{bhattacharya2010framework}, local and global features~\cite{wang2005reduced},~\etc
Representative works include, but are by no means limited to, Datta~\etal's~\cite{datta2006studying} use of K-means clustering, Marchesotti~\etal's~\cite{marchesotti2011assessing} use of Fisher vector and Bag-of-visual-words, Saad~\etal's~\cite{saad2012blind} approach with Bayesian model, 
and Support vector machines in~\cite{luo2011content,bhattacharya2013towards,mavridaki2015comprehensive}.

In summary, traditional IAA methods offer a high degree of interpretability due to their transparent feature design and prediction mechanism, but their feature representations are largely manually engineered. This limits their flexibility in modeling the rich variety of visual cues involved in aesthetic assessment.

\noindent\textbf{Observation 2: Deep learning has dramatically improved IAA performance, but at the cost of interpretability.} 
Deep learning-based approaches for IAA leverage deep neural networks to automatically learn aesthetic representations from large-scale image datasets. These models achieve remarkable performance gains because they can capture complex visual patterns that are difficult to model manually. As a result, deep learning has become the dominant paradigm in contemporary IAA research.

Existing approaches can be broadly categorized into two schemes. The first adopts generic deep features from models pretrained on classification tasks, as seen in work like~\cite{dong2015photo}.
The second and most prevalent direction involves task-specific learning from aesthetic data. This includes fine-tuning pre-trained models~\cite{bianco2016predicting,wang2016finetuning}, or employing multi-tasks learning strategies that incorporate scene and texture~\cite{kao2016hierarchical}, semantic information~\cite{kao2017deep,shi2024semantic}, 
and content-aware modules based on style attributes~\cite{lu2015rating}, content category~\cite{kong2016photo}, composition~\cite{mai2016composition}, layout~\cite{she2021hierarchical}, and more.
Other strategies, such as deep metric learning, have been used in~\cite{schwarz2018will}. 
In terms of network architecture, there is a wide variety of choices, ranging from single-column CNN~\cite{wang2016multi,tian2015query,chen2024topiq} and multi-column CNNs~\cite{lu2014rapid,wang2016brain,ren2017personalized,he2022rethinking}, to more recent Transformer-based architectures~\cite{ke2021musiq,he2023thinking,he2023eat,liu2024elta,xu2024boosting,behrad2025charm},
and Multimodal frameworks~\cite{ke2023vila,zhong2025rethinking,sheng2023aesclip,wu2024qalign,huang2025multimodality,zhou2025gamma,AesBench,liao2025humanaesexpert,jin2024apddv2}.

However, deep neural networks are inherently opaque: their predictions emerge from complex representations, making it difficult to trace the rationale behind their judgments. This has motivated a growing interest in methods that not only predict aesthetic quality accurately, but also explain why an image is perceived as pleasing or not.

\begin{figure*}[t]
  \centering
  \includegraphics[width=\linewidth]{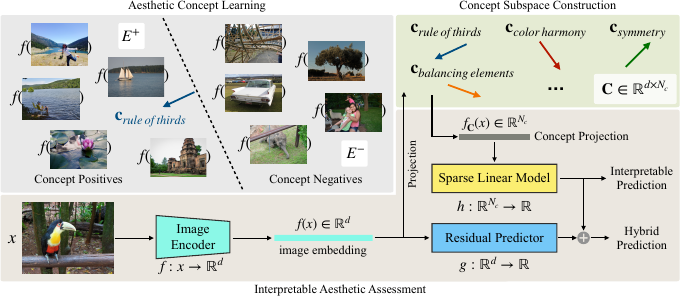}
  \vspace{-7.5mm}
  \caption{Methods overview. We predict the aesthetic score of an image using human-understandable concepts and an inherently interpretable sparse linear model.
  \textbf{(a) Aesthetic Concept Learning:} 
   Given an aesthetic concept $s$ (\textit{e.g.,} rule of thirds) and user-defined positive/negative image sets, we learn its Concept Activation Vector (CAV) $\mathbf{c}_s\in\mathbb{R}^d$ by training a linear Support Vector Machine (SVM) to distinguish between image embeddings from the two sets. The learned CAV is orthogonal to the decision boundary. 
   \textbf{(b) Concept Subspace Construction:}
   We aggregate the CAVs of $N_c$ aesthetic concepts to form a concept subspace $\mathbf{C}\in\mathbb{R}^{d \times N_c}$.
   \textbf{(c) Interpretable Aesthetic Assessment:}
   For a given image, we extract its embedding using a pre-trained image encoder and project it onto the concept subspace. The resulting concept projection is used by an inherently interpretable sparse linear model to predict the aesthetic score. To account for nuanced influences on aesthetic judgment beyond explicit concepts, we add a residual predictor that complements the interpretable core.
   }
   \label{fig:methods_overview}
\end{figure*}

\noindent{\bf Observation 3: Efforts to explain IAA models remain limited and indirect.}
There have been several attempts to improve the explainability of deep aesthetic models, but they are still relatively limited and often indirect. A common strategy is the use of visualization techniques. For instance, Sartori~\etal~\cite{sartori2015affective} applied back-projection to highlight pixel-wise contributions, while Reddy~\etal~\cite{viswanatha2020measuring} introduced attribute activation maps, other approaches include saliency maps~\cite{rubio2021application} and occlusion attribution maps~\cite{schultze2023explaining}.

Another line of work seeks to enhance explainability by generating textual critiques through multimodal models~\cite{li2024towards,huang2024aesexpert,wang2025mllms,qi2025photographer,jiang2025multimodal}, allowing users to receive natural language feedback on aesthetic judgments.

In addition, general-purpose Explainable AI techniques have been explored. Soydaner and Wagemans~\cite{soydaner2024unveiling} employed SHAP (Shapley Additive Explanations)~\cite{lundberg2017unified} to identify how aesthetic attributes contribute to overall aesthetic scores. More recently, Viriyavisuthisakul~\etal~\cite{viriyavisuthisakul2025explainable} combined SHAP with a language model to generate linguistic explanations. Outside the IAA domain, Kim~\etal~\cite{kim2018interpretability} introduced concept activation vectors to probe black-box models using directional derivatives.
However, a series of works by Rudin~\etal~\cite{rudin2019we,rudin2019stop}
argued that interpretable models should be preferred whenever possible, rather than `explained' black-box models. Inspired by this perspective, we employ an inherently interpretable sparse linear model, and directly integrate learned aesthetic concept subspaces into the prediction process.

\noindent{\bf Observation 4: Aesthetic assessments can benefit from human-nameable visual attributes.}
Human-nameable visual attributes can benefit various computer vision tasks, including IAA.
Early efforts such as Dhar~\etal~\cite{dhar2011high} proposed using human-describable attributes related to composition, content, and illumination.
One of the most successful examples is by Marchesotti~\etal~\cite{marchesotti2015discovering}, which mined textual comments to discover discriminative aesthetic attributes and associate them with visual features. For clarity, their approach~\cite{marchesotti2015discovering} relies on generic descriptors (SIFT~\cite{lowe1999object} and color statistics).
In the deep learning era, several methods have incorporated aesthetic attributes~\cite{li2023theme,pan2019image,jin2019aesthetic,huang2024predicting,kong2016photo,schultze2023explaining,jin2024paintings,zhu2023personalized}, but these are typically used as auxiliary signals to regularize network training. By contrast, our method learns aesthetic-related visual concepts directly from raw images, and grounds the final predictions explicitly in human-understandable aesthetic dimensions.

\noindent\textbf{Summary:} 
These four observations highlight the gaps in existing IAA literature that our work addresses. Our approach learns human-understandable aesthetic concepts as the foundation of aesthetic assessment, and employs an inherently interpretable model for transparent prediction. To capture nuanced factors influencing aesthetic judgment, we introduce a residual module that enhances predictive performance while maintaining interpretability. This perspective has received limited attention in prior IAA research.
\section{Methods}
\label{sec:method}

In this section, we present our method for interpretable IAA in detail. 
We consider a supervised setting where each training image $x\in\mathbb{R}^{W \times H \times 3}$ with width $W$ and height $H$ is associated with an aesthetic score $y\in\mathbb{R}$.
Our objective is to learn a model that estimates the aesthetic quality of an image accurately while remaining interpretable to humans.

We approach this goal through three components:
(1) \textbf{Aesthetic Concepts Learning:} We learn a set of human-understandable aesthetic concepts that serve as the basis for our interpretation.
(2) \textbf{Interpretable Prediction:} We employ an inherently interpretable sparse linear model. That is, the prediction is a weighted sum of comprehensible concept scores.
(3) \textbf{Residual Prediction Module:} A residual predictor captures nuanced influences beyond explicit aesthetic concepts, complementing the concept-based model to enhance predictive performance while preserving the interpretable core.
An overview of the method is illustrated in~\cref{fig:methods_overview}.

\subsection{Learning Aesthetic Concepts}
\label{sec:concepts_learning}

The first step of our method is to learn aesthetic concepts. We achieved this by learning concept activation vectors.

Given an aesthetic concept $s$, 
we first construct two datasets: a positive concept set $\mathcal{P}_s^+$ and a negative concept set $\mathcal{P}_s^-$.
For example, as illustrated in~\cref{fig:methods_overview}, for the concept `rule of thirds',
a common composition guideline,
we collect images following this rule as the positive set,
and images that do not as the negative set.
We then encode all images from both sets using an image encoder $f:x\rightarrow\mathbb{R}^d$, obtaining their corresponding embedding sets:
\begin{equation}
\label{eq:concept_embsets}
E_s^{\sigma} = \{ f(x) \in \mathbb{R}^d \mid x \in \mathcal{P}_s^{\sigma} \}, 
\quad \sigma \in \{+, -\}.
\end{equation}
In our implementation, we use CLIP-ResNet50~\cite{radford2021learning} as the image encoder $f$, where each embedding $f(x)$ is a 1024-dimensional vector.

Next, we train a binary classifier to separate positive embeddings $E_s^+$ from negative ones $E_s^-$. In practice, we use a linear-kernel SVM as the binary classifier.
The coefficient vector $\mathbf{c}_s\in\mathbb{R}^d$ of the trained classifier is taken as the Concept Activation Vector (CAV), which is orthogonal to the decision boundary (a hyperplane) and points toward the side associated with the concept positives, as indicated by the blue arrow in~\cref{fig:methods_overview}. 
Learning each concept requires only a small number of examples, and our experiments (\cref{sec:implementation_details}) show that as few as $100$ samples are sufficient to learn a robust CAV.

Since aesthetic judgments often involve multiple concepts, we construct a concept subspace by aggregating the activation vectors of individual concepts:
\begin{equation}\label{eq:concept_subspace}
    \mathbf{C} = \left\{ \mathbf{c}_i \in \mathbb{R}^d \,\middle|\, i = 1, 2, \dots, N_c \right\},
\end{equation}
where $\mathbf{c}_i$ is the activation vector of concept $i$, $N_c$ is the total number of concepts.
We treat $\mathbf{C}$ as a matrix in $\mathbb{R}^{d \times N_c}$, and define the concept subspace as: 
$\mathcal{S}_C = \operatorname{span}(\mathbf{C})$.

This approach to learning aesthetic concepts and constructing a concept subspace offers multiple advantages:
(1) \textit{Accessibility:} 
Users can define concepts using positive and negative images without needing machine learning expertise, allowing domain experts such as artists and art historians to easily participate.
(2) \textit{Customizability}: 
Users can explore a wide range of concepts that can be visually demonstrated. This is particularly suitable for aesthetic notions, which are often abstract but can be intuitively conveyed through illustrative image examples.

\subsection{Interpretable Aesthetic Assessment}
\label{sec:interpretable_pred}

After constructing the concept subspace, we project input images into this subspace and make aesthetic predictions based on their concept-level representations.

Given an input image $x$, we first use the same pre-trained image encoder $f$ from~\cref{sec:concepts_learning} to obtain its image embedding $f(x)\in \mathbb{R}^d$.
We then project $f(x)$ onto the concept subspace, resulting in the concept projection:~$f_\mathbf{C}(x)\in\mathbb{R}^{N_c}$. The $i$-th element of $f_\mathbf{C}(x)$ is computed as:
\begin{equation}\label{eq:concept_projection}
    f_\mathbf{C}^i(x)= \frac{\left< f(x), \mathbf{c}_i \right>}{\left\Vert\mathbf{c}_i\right\Vert_2^2},
\end{equation}
where $\mathbf{c}_i$ is the $i$-th concept activation vector from~\cref{eq:concept_subspace}. This projection value, $f_\mathbf{C}^i(x)$, measures the alignment between the image and concept $i$.

Next, we train a model $h:\mathbb{R}^{N_c}\rightarrow\mathbb{R}$ to predict the aesthetic score of $x$ based on its concept projection $f_\mathbf{C}(x)$.
For interpretability, we adopt a sparse linear model: a class of inherently interpretable models that learn linear relationships between input features and the output.
Let $h$ be parameterized by weights $\mathbf{w}\in\mathbb{R}^{N_c}$ and bias $b\in\mathbb{R}$. The prediction is given by:
\begin{equation}\label{eq:sparse_linear_out}
    h(f_\mathbf{C}(x))=\mathbf{w}^{\top}f_\mathbf{C}(x)+b.
\end{equation}
Here, the model computes the aesthetic score as a weighted sum of concept responses, where each weight explicitly reflects the importance of its corresponding concept. 
As noted in~\cref{sec:intro}, this design is inspired by Iigaya~\etal~\cite{iigaya2021aesthetic,iigaya2023neural}, 
who showed that human aesthetic preferences can be approximated by a linear feature summation (LFS) model through behavioral and neuroimaging experiments.

From \cref{eq:sparse_linear_out}, we can see that model $h$ provides two levels of information: (1) \textit{General aesthetic tendencies}: The learned weights $\mathbf{w}$ reflect overall aesthetic preferences in the dataset. (2) \textit{Image-specific characteristics}: The concept projection $f_\mathbf{C}(x)$ indicates how each image expresses these aesthetic concepts.
Together, these components provide transparent, human-understandable interpretation of aesthetic assessments at both the dataset and image levels.

We learn the parameters $\mathbf{w}$ and $b$ by minimizing:
\begin{equation}\label{eq:interpretable_predictor}
\scalebox{0.92}{
    $(\hat{\mathbf{w}},\hat{b})= \underset{\mathbf{w}, b}{\arg\min}       \underset{(x,y)\in\mathcal{D}}{\mathbb{E}}(\mathbf{w}^{\top}f_\mathbf{C}(x)+b-y)^2+\lambda\mathcal{R}(\mathbf{w}),$}
\end{equation}
where $\mathcal{D}$ is the training data, and $(x,y)$ denotes an image-score pair. The term $\mathcal{R}(\mathbf{w})$ is a regularization term with strength $\lambda$. We adopt Elastic Net regularization, which combines both $\ell_1$ and $\ell_2$ penalties: $\mathcal{R}(\mathbf{w})=\alpha\|\mathbf{w}\|_1+(1-\alpha)\|\mathbf{w}\|_2^2$
where $\alpha\in[0,1]$.

\subsection{Capturing Nuances Beyond Explicit Concepts}
\label{sec:residual_learning}

Experiments (\cref{sec:experiments}) demonstrate that the interpretable predictor $h$ alone provides strong predictive performance. In machine learning, the well-known Interpretability-Performance trade-off is often discussed.
In our experiments on IAA datasets, we observe that a small gap between interpretability and predictive performance remains.
This is largely because aesthetic perception involves diverse and nuanced cues, many of which extend beyond the scope of explicit, interpretable concepts.
In our framework, we prioritize interpretability by limiting the number of aesthetic concepts.
This keeps interpretations clear and not overwhelming, while some latent factors or unmodeled concepts may not be captured.
Additionally, as noted earlier, some hidden influences in aesthetic judgment are inherently difficult to formalize explicitly.
To model this remaining complexity, we introduce a residual predictor that complements the concept-based model. This predictor captures additional variance while leaving the interpretable core unchanged (as illustrated in~\cref{fig:methods_overview}).

Let $g(\mathbf{z}; \boldsymbol{\theta})$ be the residual predictor, where $\mathbf{z} = f(x)$ is the image embedding and $\boldsymbol{\theta}$ represents the model parameters.
We learn $\boldsymbol{\theta}$ by solving the following optimization problem:
\begin{equation}\label{eq:residual_predictor_optimize}
    \hat{\boldsymbol{\theta}}= \arg\underset{\boldsymbol{\theta}}{\min}       \underset{(x,y)\in\mathcal{D}}{\mathbb{E}}(h(f_\mathbf{C}(x))+g(f(x);\boldsymbol{\theta})-y)^2,
\end{equation}
where $(x, y)$ is an image-score pair from the training data $\mathcal{D}$, and $h(f_\mathbf{C}(x))$ is the interpretable
prediction defined in~\cref{eq:sparse_linear_out}.
To maintain simplicity, we implement $g$ as a linear regression:
\begin{equation}\label{eq:residual_predictor_output}
    g(f(x); \boldsymbol{\theta}) = \mathbf{w}_r^{\top} f(x) + b_r,
\end{equation}
where $\boldsymbol{\theta}=(\mathbf{w}_r, b_r)$ denotes the parameters of $g$.

It should be noted that residual learning is performed sequentially: both the concept subspace $\mathbf{C}$ (~\cref{eq:concept_subspace}) and the interpretable model $h$ (\cref{eq:sparse_linear_out}) are kept fixed during the training of $g$ (\cref{eq:residual_predictor_optimize}).
The residual predictor serves as a supplementary component; it enhances the overall prediction without modifying the interpretable part. This design provides a simple and effective way to capture additional variance, boosting predictive performance while preserving the clarity of the concept-based model.

\section{Experiments}
\label{sec:experiments}

In this section, we evaluate our approach with two objectives:
(1) to compare predictive performance against state-of-the-art baselines, and (2) to assess the interpretability of the resulting predictions.
Experiments are conducted on both photographic and artistic datasets, with results demonstrating competitive accuracy and clear interpretations.

\subsection{Datasets and Evaluation Metrics}\label{sec:datasets}

We evaluate our method on five Image Aesthetic Assessment (IAA) datasets, using two standard evaluation metrics.

\noindent\textbf{Photo Domain Datasets:} 
\begin{itemize}[label=\textbullet, topsep=0pt, partopsep=0pt, parsep=0pt, itemsep=0pt]
    \item AADB~\cite{kong2016photo}:
     $10$k images, each with a normalized aesthetic score in $[0,1]$ and normalized scores for eleven attributes.
     \item PARA~\cite{yang2022personalized}:
     $31{,}200$ images, each annotated with a mean aesthetic score in $[1,5]$ along with attributes.
     \item AVA~\cite{murray2012ava}: 
     $\mathord{\sim}250$k images, each annotated with a score distribution on a $1\text{--}10$ scale. 
\end{itemize}

\noindent We learn photo-related aesthetic concepts from two datasets, AADB and PARA, and use them to predict aesthetic scores on the same datasets. 
In addition, we use these learned concepts to evaluate our approach on the AVA dataset.

\noindent\textbf{Artistic Domain Datasets:}
\begin{itemize}[label=\textbullet, topsep=0pt, partopsep=0pt, parsep=0pt, itemsep=0pt]
    \item LAPIS~\cite{maerten2025lapis}: $11{,}723$ art images curated with art historians, each with an aesthetic score in $[0, 100]$ and attributes related to aesthetic appreciation.
    \item BAID~\cite{yi2023towards}: $60{,}337$ artworks with preference votes scaled to $[0, 10]$; higher scores indicate higher aesthetic value.
\end{itemize}

\noindent We learn art-related aesthetic concepts from the LAPIS dataset. The learned concepts are then used for aesthetic score prediction on both LAPIS and BAID.

\noindent\textbf{Evaluation Metrics:}
We use two commonly used metrics: (1) Spearman's Rank Correlation Coefficient (SRCC), which measures the monotonic relationship between predictions and ground truth; and (2) Pearson's Linear Correlation Coefficient (PLCC), which measures their linear relationship. Both metrics range from $-1$ to $1$, with higher values indicating stronger agreement.

\subsection{Implementation Details}\label{sec:implementation_details}

\noindent \textbf{Aesthetic Concepts for Photos:}
For the \textbf{AADB} dataset, we use the $11$ aesthetic attributes provided by the dataset creators after consultation with professional photographers:  \textit{interesting content, object emphasis,}~\etc (see~\cref{fig:aadb_concept_weights}).
Each attribute is annotated with a score in the range $[-1, 1]$,
For each attribute $s$, we select the top $100$ images as the positive set $\mathcal{P}_s^+$, and the bottom $100$ images as the negative set $\mathcal{P}_s^-$,
and use these sets to learn the corresponding aesthetic concept vector, as described in~\cref{sec:concepts_learning}.
For the \textbf{PARA} dataset, we follow a similar procedure. The only difference lies in the number and type of aesthetic attributes: we use $9$ attributes provided by the dataset creator that partially overlap with those of AADB (see \cref{fig:para_concept_weights}).

\noindent \textbf{Aesthetic Concepts for Art Images:}
The aesthetic concepts for art images differ from those for photographs. Artworks are generally created by professional artists and often follow foundational aesthetic principles. For observers, aesthetic judgments are shaped by many factors, with psychology researches~\cite{cela2002style,belke2006mastering} showing that styles and genres play a particularly prominent role. The \textbf{LAPIS} dataset provides annotations for $26$ styles (ranging
from \textit{Renaissance} to \textit{Minimalism}) and $7$ genres (\textit{abstract,
cityscape, flower painting, landscape, nude painting, portrait,} and \textit{still life}).
We construct the concept subspace for art images using activation vectors derived from these $33$ attributes. 
For this art dataset, style and genre annotations are binary rather than continuous. Therefore, positive and negative sets are constructed via random sampling instead of ranking-based selection. Specifically:
\begin{itemize}[label=\textbullet, topsep=0pt, partopsep=0pt, parsep=0pt, itemsep=0pt]
    \item For each of the $26$ styles, we randomly select $200$ images as the positive set, and sample $8$ images from each of the remaining 25 styles ($8\times25=200$) as the negative set.
    \item For each of the $7$ genres, we randomly select $150$ images as the positive set, and sample $25$ images from each of the remaining $6$ genres ($25\times6=150$) as the negative set.
\end{itemize}

\begin{table}[t]
  \centering
  \renewcommand{\arraystretch}{1.06}
  \begin{tabular}{>{\hskip 1pt}l<{\hskip 20pt}c<{\hskip 40pt}c<{\hskip 1pt}}
    \toprule
    Methods & SRCC & PLCC  \\
    \midrule
    NIMA~\cite{talebi2018nima}  & 0.708 & 0.711 \\
    MLSP~\cite{hosu2019effective} & 0.725 & 0.726 \\
    RGNet~\cite{liu2020composition} & 0.710 & - \\
    PA-IAA~\cite{li2020personality} & 0.720 & 0.728  \\
    MUSIQ~\cite{ke2021musiq} & 0.706 & 0.712  \\
    DINOv2-small~\cite{oquab2024dinov2} & 0.682 & 0.695  \\
    Charm~\cite{behrad2025charm} & \textbf{0.754} & \textbf{0.767} \\
    CLIP-ResNet50~\cite{radford2021learning} & 0.740 & 0.745  \\
    \midrule
    \rowcolor{gray!10} 
    Ours (interpretable) & 0.697 & 0.701  \\
    \rowcolor{gray!10} 
    Ours (hybrid) & \underline{0.745} & \underline{0.748} \\
    \bottomrule
  \end{tabular}
  \vspace{-2.5mm}
  \caption{Performance comparison on the AADB dataset.
  `-' indicates an unreported metric in the original paper.
  Best and second-best results are highlighted in bold and underlined, respectively.}
  \label{tab:aadb}
\end{table}

\subsection{Performance Comparison and Analysis}\label{sec:comparison}

We select several image aesthetic assessment methods as baselines.
Our method uses CLIP-ResNet50~\cite{radford2021learning} as the image encoder ($\mathord{\sim}38$M parameters).
The extra components consist of the concept subspace ($1024$ parameters per concept), the sparse linear module (one parameter per concept), and the residual predictor ($1024$ parameters), all of which are negligible compared to the backbone.
Therefore, we mainly compare against methods with comparable parameter scales.
These include CNN-based methods (\eg,~\cite{talebi2018nima},~\cite{hosu2019effective},~\etc),
more recent Transformer-based approaches~\cite{ke2021musiq,oquab2024dinov2},
and the state-of-the-art algorithm Charm~\cite{behrad2025charm}.
We also compare our approach with an explainable IAA method~\cite{soydaner2024unveiling}.
In addition, we apply a linear regression probe on top of CLIP-ResNet50 as a baseline. All results are based on the same
training/test split of the dataset and are either quoted directly from the original papers or obtained by running the official codes.
Reported results are averaged over three runs.

It is important to note that our focus lies in both interpretability and performance. Our aim is not merely to maximize accuracy, but to achieve strong performance while ensuring interpretability. Results demonstrate that our method achieves competitive performance compared to standard models, with the additional advantage of interpretability.

\begin{figure}[t]
  \centering
  \includegraphics[width=\linewidth]{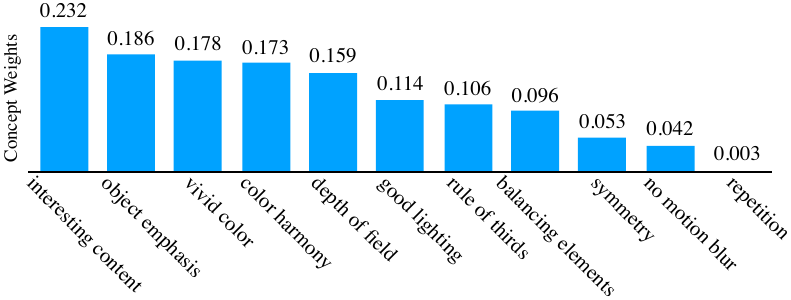}
  \vspace{-7mm}
  \caption{Learned aesthetic concept weights on the AADB dataset, ordered by importance (corresponding to $\mathbf{w}$ in \cref{eq:sparse_linear_out}). The bias term is 0.538 (corresponding to $b$ in \cref{eq:sparse_linear_out}).
   }
   \label{fig:aadb_concept_weights}
\end{figure}

\begin{figure}[b]
  \centering
  \includegraphics[width=\linewidth]{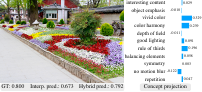}
  \vspace{-7mm}
  \caption{Aesthetic score prediction on the AADB test image.
   Bottom left shows the ground truth (GT), 
   our interpretable prediction (Interp. pred.), 
   and hybrid prediction (Hybrid pred.). 
   The right side shows the image's projection on the learned concept subspace.
   }
   \label{fig:aadb_concept_projs}
\end{figure}

\noindent\textbf{Results on AADB:}
~\cref{tab:aadb} reports the performance of different methods on the AADB dataset. Overall, both CNN-based~\cite{talebi2018nima,hosu2019effective,liu2020composition,li2020personality} and Transformer-based~\cite{ke2021musiq,oquab2024dinov2,behrad2025charm} approaches achieve comparable results.
The recently proposed state-of-the-art method, Charm~\cite{behrad2025charm}, attains the highest performance.
Our hybrid model gets the second-best results, slightly outperforming the CLIP-ResNet50 baseline.
Our interpretable branch alone scores lower, but still provides a satisfactory level of performance.

\cref{fig:aadb_concept_weights} presents the learned weights of the aesthetic concepts on AADB, ordered from highest to lowest.
All weights are positive, indicating that each concept contributes positively to the overall aesthetic score, which aligns with insights from professional consultation. 
Among the concepts, \textit{interesting content} is the most influential factor, followed by \textit{object emphasis} and \textit{color}-related attributes.
Mid-level contributions come from structural attributes, such as \textit{rule of thirds, balancing} and \textit{symmetry}. In contrast, \textit{repetition} shows only a negligible effect.
These results suggest that the model captures a general tendency of participants in this dataset to prioritize semantic and color-related features over purely structural patterns.

\cref{fig:aadb_concept_projs} illustrates our model's predictions on a sample test image from AADB. The interpretable model produces a score of $0.673$, reasonably close to the ground truth $0.800$, while the hybrid model further improves precision with a prediction of $0.792$.
More importantly, the right panel shows the image's projection on the concept subspace. 
The model identifies \textit{vivid color, color harmony,} and \textit{rule of thirds} as the most prominent aesthetic attributes, which we believe aligns well with common subjective judgments of this image.  This demonstrates that our approach not only makes reliable predictions but also provides clear interpretations, making the results transparent and trustworthy.

\begin{table}[b]
  \centering
  \renewcommand{\arraystretch}{1.06}
  \begin{tabular}{>{\hskip 1pt}l<{\hskip 20pt}c<{\hskip 35pt}c<{\hskip 1pt}}
    \toprule
    Methods & SRCC & PLCC  \\
    \midrule
    NIMA~\cite{talebi2018nima}  & 0.875 & 0.862 \\
    BIAA~\cite{zhu2020personalized}  & 0.858 & 0.886 \\
    HGCN~\cite{she2021hierarchical} & 0.865 & 0.881 \\
    DINOv2-small~\cite{oquab2024dinov2} & 0.855 & \underline{0.904}  \\
    Charm~\cite{behrad2025charm} & \textbf{0.905} & \textbf{0.938} \\
    CLIP-ResNet50~\cite{radford2021learning} & 0.872 & 0.903  \\
    \midrule
    \rowcolor{gray!10} Ours (interpretable) & 0.803 & 0.828  \\
    \rowcolor{gray!10} Ours (hybrid) & \underline{0.877} & 0.901  \\
    \bottomrule
  \end{tabular}
  \vspace{-2.5mm}
  \caption{Performance comparison on PARA.}
  \label{tab:para}
\end{table}

\noindent\textbf{Results on PARA:}
\cref{tab:para} summarizes the results on PARA. 
Overall, performance differences across methods are relatively small, with Charm achieving the highest scores on both metrics.
Our hybrid model ranks just behind, reaching competitive performance in both SRCC and PLCC.
The interpretable branch performs lower, as expected, but still provides a solid level of predictive performance.
Together, these results indicate that our method maintains a strong balance between predictive accuracy and interpretability.

\begin{figure}[t]
  \centering
  \includegraphics[width=\linewidth]{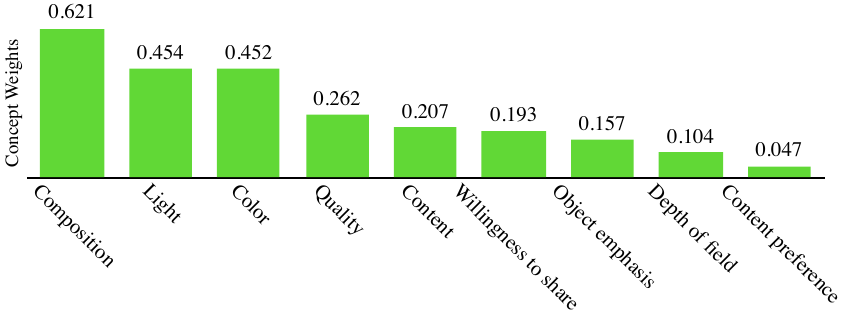}
  \vspace{-7.5mm}
   \caption{Learned weights of aesthetic concepts on the PARA dataset. The learned bias term is $3.017$. 
   }
   \label{fig:para_concept_weights}
\end{figure}

\begin{figure}[htbp]
  \centering
  \includegraphics[width=\linewidth]{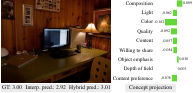}
  \vspace{-7mm}
  \caption{Aesthetic score prediction on the PARA test image. 
   Bottom left shows the ground truth and our predictions. The right side shows the image's projection on the learned concept subspace.
   }
   \label{fig:para_concept_projs}
\end{figure}

\cref{fig:para_concept_weights} shows the learned weights of the aesthetic concepts on PARA. Unlike AADB, \textit{Composition} stands out as the dominant factor, followed closely by \textit{Light} and \textit{Color}, highlighting the importance of structural and visual elements in participants' ratings. Concepts such as \textit{Quality} and \textit{Content} have moderate contributions, while \textit{Content preference} plays only a minor role. 
Overall, the weight distribution reflects how the model captures underlying aesthetic trends in annotators' judgments for this dataset.
\cref{fig:para_concept_projs} illustrates the aesthetic score prediction for a test image. Both interpretable and hybrid predictions closely match the ground truth. The concept projection indicates that model identifies \textit{Composition} as the most prominent attribute, reflecting the well-structured arrangement of the objects. \textit{Depth of field} is also captured, while attributes such as \textit{Color} and \textit{Quality} receive negative values, consistent with the dim lighting and limited clarity. These results demonstrate that the model can accurately assess both appealing and unappealing aesthetic attributes of the image.

\noindent\textbf{Results on AVA:} We further evaluate our approach on the AVA dataset, using concepts learned from AADB and PARA. 
The hybrid model maintains competitive performance in this setting.

\begin{table}[htbp]
  \centering
  \renewcommand{\arraystretch}{1.06}
  \begin{tabular}{>{\hskip 1pt}l<{\hskip 20pt}c<{\hskip 30pt}c<{\hskip 1pt}}
    \toprule
    Methods & SRCC & PLCC  \\
    \midrule
    LAPIS~\cite{maerten2025lapis} & \underline{0.809} & \underline{0.813} \\
    \rowcolor{gray!10} Ours (interpretable) & 0.805 & 0.800  \\
    \rowcolor{gray!10} Ours (hybrid) & \textbf{0.817} & \textbf{0.815}  \\
    \bottomrule
  \end{tabular}
  \vspace{-2.5mm}
  \caption{Performance comparison on LAPIS.}
  \label{tab:lapis}
\end{table}

\begin{figure}[b]
  \centering
  \includegraphics[width=\linewidth]{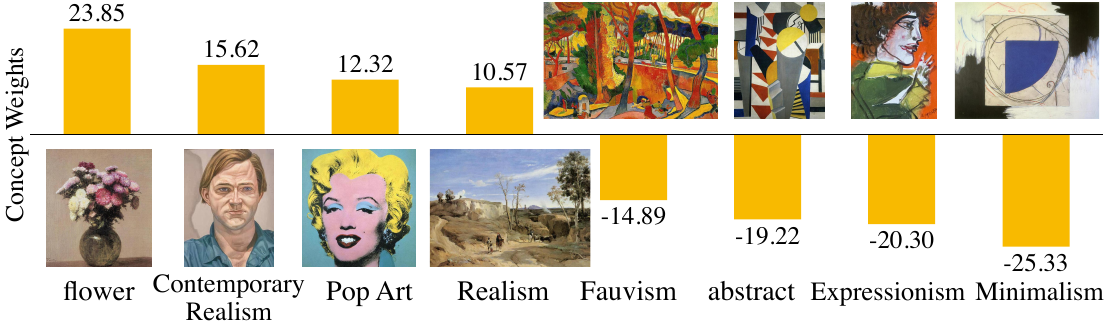}
  \vspace{-7.5mm}
  \caption{The weights of aesthetic concepts on the LAPIS dataset, we list the four with the highest and lowest weights. For easier understanding, a representative image is included for each one.
   }
   \label{fig:lapis_concept_projs}
\end{figure}

\noindent\textbf{Results on LAPIS:}
\cref{tab:lapis} presents results on the artworks dataset LAPIS. As reported in the original paper, the baseline method~\cite{maerten2025lapis} achieves solid performance.
Our interpretable branch performs slightly below this baseline, while the hybrid model achieves marginally higher scores.

\cref{fig:lapis_concept_projs} highlights the most and least influential concepts on the dataset. The \textit{flower} genre carries the highest weight, suggesting that participants tend to rate flower paintings more favorably. The next three concepts, \textit{Contemporary Realism, Pop Art,} and \textit{Realism}, all fall within representational figurative art, reflecting a general preference for visually recognizable styles. In contrast, \textit{Minimalism}, a non-representational abstract style, receives the lowest weight, with \textit{Expressionism, Fauvism,} and the \textit{abstract} genre also ranking among the lowest. These results are consistent with the dataset creators' observation that ``abstract works generally receive lower aesthetic scores than figurative ones"~\cite{maerten2025lapis}, while also demonstrating how our model captures this trend via the learned concept weights.

\begin{table}[t]
  \centering
  \renewcommand{\arraystretch}{1.06}
  \begin{tabular}{>{\hskip 1pt}l<{\hskip 20pt}c<{\hskip 35pt}c<{\hskip 1pt}}
    \toprule
    Methods & SRCC & PLCC  \\
    \midrule
    NIMA~\cite{talebi2018nima}  & 0.393 & 0.382 \\
    MPADA~\cite{sheng2018attention}  & 0.437 & 0.425 \\
    MLSP~\cite{hosu2019effective} & 0.441 & 0.430 \\
    TANet~\cite{he2022rethinking} & 0.453 & 0.437  \\
    SAAN~\cite{yi2023towards} & \underline{0.473} & \underline{0.467} \\
    DINOv2-small~\cite{oquab2024dinov2} & 0.342 & 0.428  \\
    Charm~\cite{behrad2025charm} & 0.368 & 0.439 \\
    CLIP-ResNet50~\cite{radford2021learning} & 0.435 & 0.443  \\
    \midrule
    \rowcolor{gray!10} Ours (interpretable) & 0.380 & 0.340  \\
    \rowcolor{gray!10} Ours (hybrid) & \textbf{0.502} & \textbf{0.482}  \\
    \bottomrule
  \end{tabular}
  \vspace{-2.5mm}
  \caption{Performance comparison on BAID.}
  \label{tab:baid}
\end{table}

\begin{figure}[b]
  \centering
  \includegraphics[width=\linewidth]{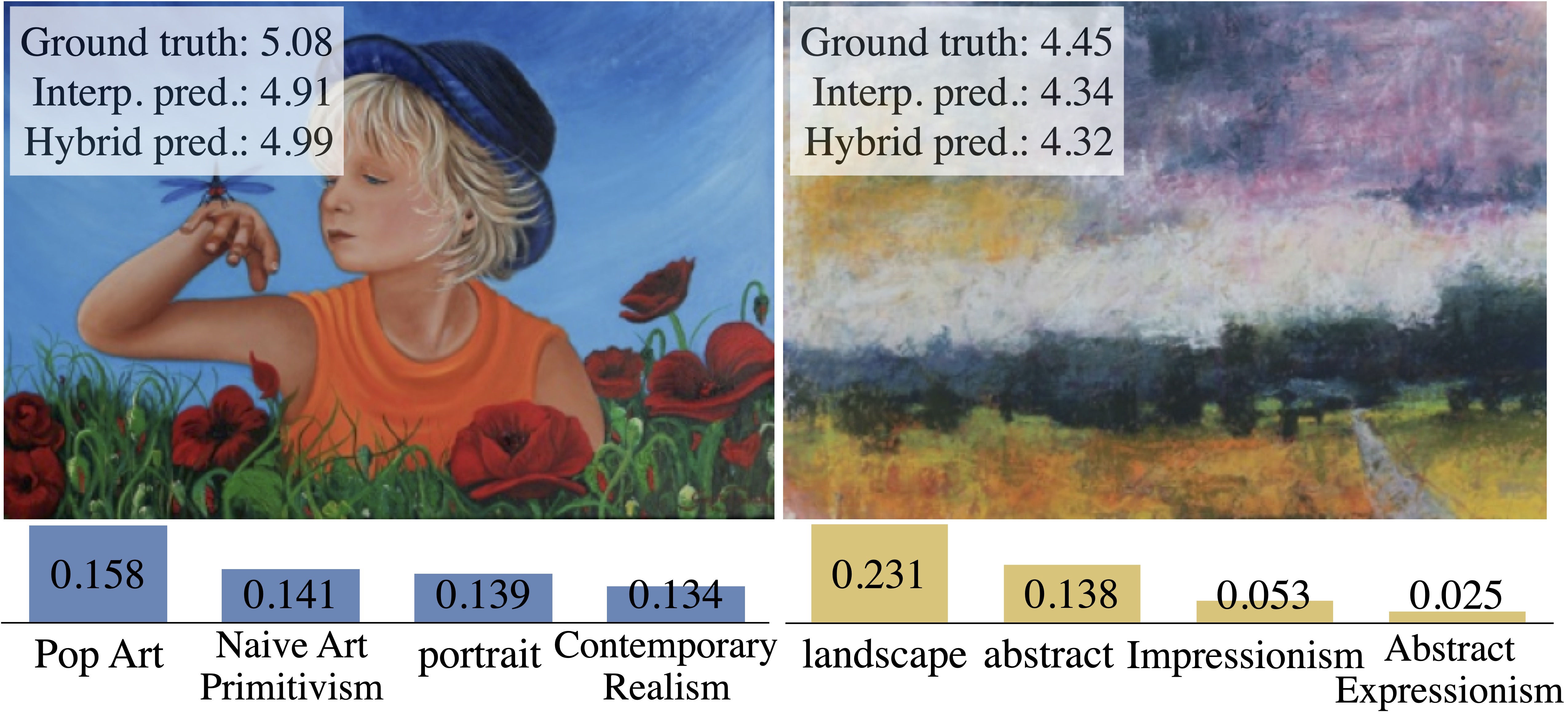}
  \vspace{-7.5mm}
  \caption{Aesthetic score predictions for BAID test images. The bottom shows the four highest concept projection values and their corresponding concept names for each image.
   }
   \label{fig:baid_test}
\end{figure}

\noindent\textbf{Results on BAID:}
We apply the concept subspace for art images, constructed from LAPIS, to BAID. The results are reported in \cref{tab:baid}. Overall, all methods achieve relatively modest performance. One important reason lies in the nature of BAID: the images were sourced from the BoldBrush platform, where monthly contests allow users to vote for artworks they like. These contests typically focus on specific themes. The votes were then converted into aesthetic scores. 
Consequently, unlike conventional IAA datasets which are based on explicit aesthetic ratings, BAID's labels largely reflect thematic fit and popularity in a contest setting, and may not fully represent aesthetic judgments.

This scenario highlights the advantages of our interpretable framework, as it allows clear tracing of how predictions are formed.
While our hybrid model achieves the best overall performance on BAID, the interpretable branch provides valuable insights beyond predictive accuracy.
The learned bias is $4.044$, with \textit{Abstract Expressionism} receiving the highest weight ($0.473$), followed by \textit{Naive Art Primitivism} ($0.269$), while \textit{Fauvism} obtains the lowest ($-0.665$). \cref{fig:baid_test} shows predictions for two test images. The concept projections clearly capture the stylistic essence of each artwork, for example, the portrait on the left and the abstract landscape on the right. Combined with the learned bias and concept weights, these projections provide a transparent interpretation of how the final scores are produced.

\begin{figure}[b]
  \centering
  \includegraphics[width=0.95\linewidth]{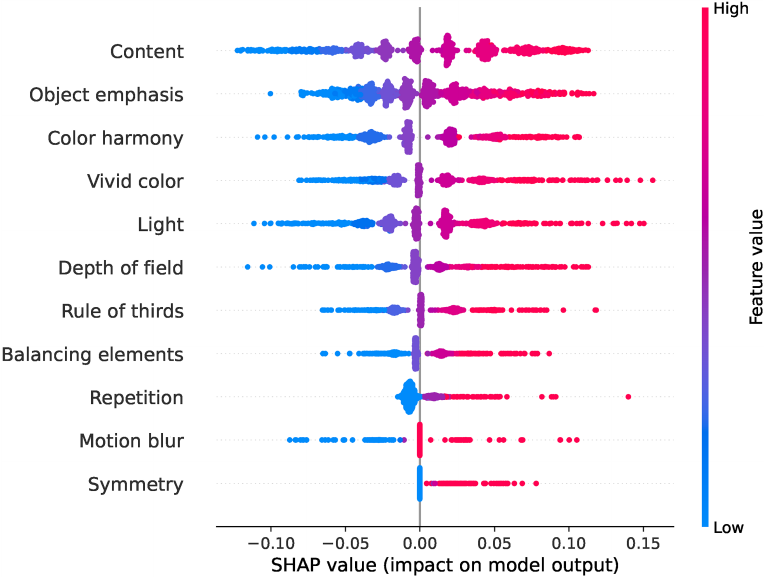}
  \vspace{-3mm}
  \caption{Attribute importance ranking on the AADB dataset obtained by Explainable IAA~\cite{soydaner2024unveiling}.
   }
   \label{fig:compare_explainiaa}
\end{figure}

\noindent\textbf{Comparison with Explainable IAA:}
To further validate the interpretability of our method, we compare it with an explainable IAA approach~\cite{soydaner2024unveiling}.
This approach employs the Explainable AI technique SHAP~\cite{lundberg2017unified} to provide post-hoc explanations of trained models, producing a ranking of feature importance.

\cref{fig:compare_explainiaa} shows the SHAP-based attribute importance ranking produced by Explainable IAA~\cite{soydaner2024unveiling} on the AADB dataset~\cite{kong2016photo}, with the most influential at the top.
The attribute \textit{content} emerges as the most important, followed by \textit{object emphasis} and \textit{color harmony}, while \textit{repetition, motion blur,} and \textit{symmetry} have minimal impact.

Notably, the ranking of our learned aesthetic concept weights on AADB (\cref{fig:aadb_concept_weights}) is closely aligned with that produced by Explainable IAA: \textit{content} is most influential, followed by \textit{object emphasis} and \textit{color}-related attributes, while \textit{symmetry, motion blur,} and \textit{repetition} are least influential.

This demonstrates that both approaches capture similar human-perceived priorities. Importantly, our framework achieves this interpretability inherently, without relying on post-hoc analysis.

\subsection{Limitations and Future Work}
\label{sec:limitation}

Our method employs a linear formulation for the interpretable predictor, which can limit its ability to capture complex interactions between aesthetic concepts.
In practice, certain aesthetic effects may arise from synergistic or context-dependent combinations of concepts.
Future work could explore more expressive yet still human-understandable interactions among aesthetic concepts, or progressively refine and personalize concept sets to better reflect diverse aesthetic judgments. Insights from empirical aesthetics could help guide these directions.

\section{Conclusion}

In this paper, we introduce a new modeling perspective for interpretable image aesthetic assessment. 
Our framework learns human-understandable aesthetic concepts 
and organizes them into a concept subspace that serves as the basis of an inherently interpretable model.
To capture nuanced influences beyond explicit concepts, we incorporate a residual predictor that complements the interpretable core without affecting its transparency. Experimental results demonstrate that our method achieves competitive performance across diverse datasets in both photography and art domains, while providing meaningful interpretive insights. 

It is clear that the gulf between machine and human aesthetic judgments remains a strong motivation for future research. Leveraging human-understandable concepts within interpretable models opens many promising avenues.

{
  \small
  \bibliographystyle{ieeenat_fullname}
  \bibliography{main}
}

\end{document}